\documentclass[a4paper,twoside]{article}

\usepackage{epsfig}
\usepackage{subcaption}
\usepackage{calc}
\usepackage{amssymb}
\usepackage{amstext}
\usepackage{amsmath}
\usepackage{amsthm}
\usepackage{multicol}
\usepackage{pslatex}
\usepackage{apalike}
\usepackage{SCITEPRESS}     % Please add other packages that you may need BEFORE the SCITEPRESS.sty package.

\begin{document}

\title{Semi-Supervised Surface Anomaly Detection of Composite Wind Turbine Blades From Drone Imagery}

% Semi-Supervised End-to-End Surface Anomaly Detection of Composite Wind Turbine Blades From Drone Imagery

\author{\authorname{Jack. W. Barker \sup{1}, Neelanjan Bhowmik \sup{1} and Toby. P. Breckon\sup{1, 2}}

\affiliation{Department of \{Computer Science \sup{1} $|$ Engineering \sup{2}\}, Durham University, Durham, UK}
}

\abstract{Within commercial wind energy generation, the monitoring and predictive maintenance of wind turbine blades in-situ is a crucial task, for which remote monitoring via aerial survey from an Unmanned Aerial Vehicle (UAV) is commonplace. Turbine blades are susceptible to both operational and weather-based damage over time, reducing the energy efficiency output of turbines. In this study, we address automating the otherwise time-consuming task of both blade detection and extraction, together with fault detection within UAV-captured turbine blade inspection imagery. We propose BladeNet, an application-based, robust dual architecture to perform both unsupervised turbine blade detection and extraction, followed by super-pixel generation using the Simple Linear Iterative Clustering (SLIC) method to produce regional clusters. These clusters are then processed by a suite of semi-supervised detection methods. Our dual architecture detects surface faults of glass fibre composite material blades with high aptitude while requiring minimal prior manual image annotation. BladeNet produces an Average Precision (AP) of $0.995$ across our Ørsted blade inspection dataset for offshore wind turbines and 0.223 across the Danish Technical University (DTU) NordTank turbine blade inspection dataset. BladeNet also obtains an AUC of $0.639$ for surface anomaly detection across the Ørsted blade inspection dataset.}

\onecolumn \maketitle \normalsize \setcounter{footnote}{0} \vfill
\section{Introduction} \label{sec:introduction}
 Global energy demand is increasing significantly. Between 1971 to 2010, demand for energy increased $2.4$ fold ($+134$\%) and is predicted to increase by $+204.2$\% by the year 2030 \cite{Matsuo2013}. The `1992 - Kyoto Protocol', introduced by the United Nations Framework Convention on Climate Change (UNFCCC), entered into force in 2005. The Kyoto Protocol regulates 192 member countries to limit and reduce Greenhouse Gas (GHG) emissions in line with agreed individual targets. 
 
 Renewable energy sources emit negligible CO$_{2}$ emissions and can supply for the increase in demand for power. The Global Wind Energy Council (GWEC) estimates a $17$-fold increase in wind power generation, providing as much as $25 - 30$\% of global electricity by the year 2050 \cite{GWEC}, equating to $123$ petawatt-hours (PWh) of electricity annually \cite{Archer2005}. 
 
   \begin{figure}[htb!]
\centering
\includegraphics[width=\linewidth]{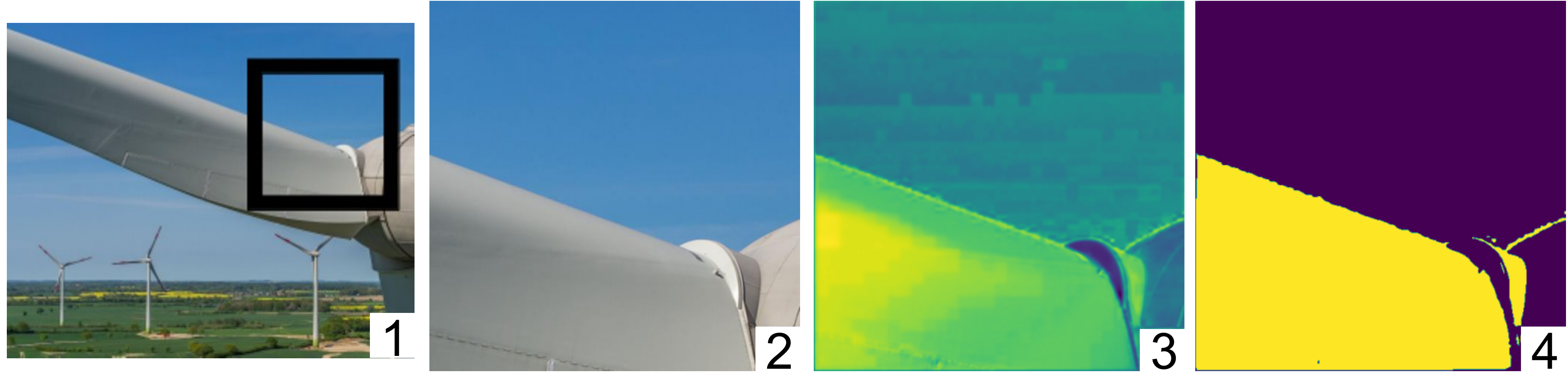}
\caption{Transfer detection of an out-of-dataset turbine blade illustrating the robust ability of our method 1) Image of wind turbine with marked region on the blade and nacelle, 2) Cropped region of turbine blade, 3) Raw model output, 4) Threshold model output producing final blade detection. }
\label{fig:blade_detections}
\end{figure}

 Unlike the reliability of fossil fuel-based energy sources to produce energy on demand however, wind energy is temperamental. Low wind speeds do not provide sufficient lift forces for turbine blades to rotate whereas high wind speeds exceeding $>25m/s$ ($90 km/h$), commonly force many modern turbines to shut down as a safety measure \cite{Sinden2007}.
 
 Few locations provide reliable and sufficient supply of wind to meet energy demands. Offshore wind farms are now favoured due to factors which include: the availability of large continuous areas suitable to major projects, and the reduction of visual or noise impact. This promotes construction of broad, widespread wind farms featuring multitudinous, larger turbines at offshore sites which generate significantly more power than their smaller, onshore counterparts. An example as to the scale of modern offshore wind farms is the Hornsea 1 wind farm which contains $174$ turbines spread across an area of 407km$^2$. Due to exhaustive usage and weather-related degradation, turbines must be routinely inspected for damage. A common cause of failure is turbine blade damage such as: erosion, kinetic foreign object collision, lightning or other weather related phenomenon, and delamination to name only a few.

 Wind  Turbine  Blades  are  typically  made from fibre-reinforced composites due to such materials exhibiting heterogeneous \cite{Leon2017} and anisotropic properties \cite{meng2020}. Typically they are constructed from Glass Fibre Reinforced Plastic (GFRP) materials \cite{Leon2017}.

 GFRP offers the material properties of being both strong (able to withstand an applied stress without failure), and ductile (able to stretch without snapping). These properties are desirable for wind turbine blades due to the strain of operational forces (constant torque forces from lift and rotation) as well as natural forces from weather fronts and foreign object collision during operation. Over time, these forces can cause damage to the blades which may require a turbine to halt operation for a period of time, or even necessitate operational cessation of the turbine, which are both costly. This is why they must be routinely and regularly checked to prevent such events \cite{Juengert2009}. In the example of the Hornsea 1 farm, each turbine on the farm has 3 blades equating to 522 total blades each with an approximate surface area of 600 $m^{2}$. Due to the sheer area, quantity, and size of turbines in new offshore wind farms, engineers and inspectors experience tremendous strain to inspect turbine blades for damage to prevent costly failures.

 In this work, we propose BladeNet, a dual module Convolutional Neural Network (CNN) architecture tool for detecting surface faults in wind turbines while requiring minimal annotation or human intervention during training. BladeNet operates in two stages: 

\begin{enumerate}
    \item Unsupervised blade detection and extraction: This allows us to remove cluttered background in a given image by a produced instance segmentation mask of the blade.
    \item Semi-supervised anomaly detection over superpixels of the detected blades: To detect anomalous regions on the surface of the blades.
\end{enumerate}

As a result, a trained engineer can evaluate the health of the wind turbine blade by observing anomalous blade regions flagged by the anomaly detection module of BladeNet.     
 
%  1) Unsupervised blade detection and extraction - This allows us to remove cluttered background in a given image by a produced instance segmentation mask of the blade. 2) Semi-supervised anomaly detection over superpixels of the detected blades - To detect anomalous regions on the surface of the blades. From this, a trained engineer can evaluate the health of the wind turbine blade by observing anomalous sub-sections flagged by BladeNet of given blades.

\section{Related Work} \label{sec:proposal}

Prior work is considered over three primary areas of focus for this work: object detection (Section \ref{subsec:object_detection}), semi-supervised anomaly detection (Section \ref{subsec:Semi-supervised_AD}) and detection of surface faults in wind turbine blades (Section \ref{subsec:fault_detection_WTB}). 

\subsection{Object Detection} \label{subsec:object_detection}
Object detection is the task of recognising, classifying and localising instances of one or many objects in images. Dominating this field are two contemporary families of approaches: Region-Based Convolutional Neural Network (R-CNN) \cite{RCNN-Girshick2014,FastRCNNRoss2015,FasterRCNNRen2015,MaskRCNNHe2017} and You Only Look Once (YOLO) \cite{YOLORedmon2016,YOLO9000Redmon2017,YOLOv3Redmon2018}.

The work of \cite{RCNN-Girshick2014} introduces the usage of CNN for object detection with the R-CNN method, in which selective search is used to extract 2000 region proposals from an image which are then individually classified using CNN features and a Support Vector Machine layer. R-CNN exhibits long inference time due to the large amount of region proposals, meaning that R-CNN cannot be used for real-time applications. Fast-R-CNN \cite{FastRCNNRoss2015} is proposed to combat this by generating convolutional feature maps of an image and identify Regions Of Interest (ROI). ROI pooling is used to reshape them into a fixed size to be classified and refined. Faster R-CNN \cite{FasterRCNNRen2015} further improves inference time by replacing selective search with a Region Proposal Network (RPN). 

Mask R-CNN \cite{MaskRCNNHe2017} is a method which introduces the notion of producing high-quality segmentation masks for detected object instances. It extends Faster R-CNN \cite{FasterRCNNRen2015} by adding a mask prediction branch in parallel with the existing branch for bounding box recognition. While Mask R-CNN \cite{MaskRCNNHe2017} can capture instance segmentation well, it is limited by a static threshold on the Intersection over Union (IoU). To address this, Cascade Mask R-CNN \cite{CascadeMaskRCNNCai2018} implements a set of sequentially trained detectors each with increasing IoU threshold value.  

In the work of \cite{YOLORedmon2016}, a one stage detector architecture, YOLO is proposed. One limitation of the R-CNN family are that they concentrate solely on image parts with high probability of containing objects whereas YOLO considers the entire image. In YOLO, the image is first split into $n \times n$ grid squares and for each grid, YOLO predicts the bounding box and their respective classification for objects. YOLO9000 \cite{YOLO9000Redmon2017} applies vast improvements to the original YOLO architecture including using direct location prediction to bound location using logistic activation, a 19-layer backbone, batch normalisation, k-means clustering over IoU and WordTree which aggregates object class labels with ImageNet labels using a hierarchical WordNet \cite{WordNetMiller1995}. Furthermore, YOLOv3 \cite{YOLOv3Redmon2018} builds on YOLO9000 with the use of a new 53-layer backbone that utilises residual connections, as well as improvements to the bounding box prediction step and a Feature Pyramid Scheme \cite{FPNLin2017} of feature extraction. 

The more recent work of YOLACT (You Only Look At CoefficienTs) \cite{YOLACTBolya2019} is most similar to our proposed model, BladeNet, implementing a fully convolutional model for real-time instance segmentation. YOLACT++ \cite{YOLACT++Bolya2020} speeds up performance by breaking the instance segmentation task into two parallel, independent sub-tasks of: generating sets of prototype masks and predicting per-instance mask coefficients respectively. 

In this work, BladeNet utilises a one-class fully convolutional architecture, based on U-NET \cite{UNetRonneberger2015} which implements skip-connections between early features in the encoder with de-convolutional, up-sampling layers in the decoder to carry information forward in the architecture. The up-sampling from latent representation to image space allows the production of high-resolution instance segmentation mask which captures detailed and sharp edges at pixel-level.

 \begin{figure*}[htb!]
\centering
\includegraphics[width=\linewidth]{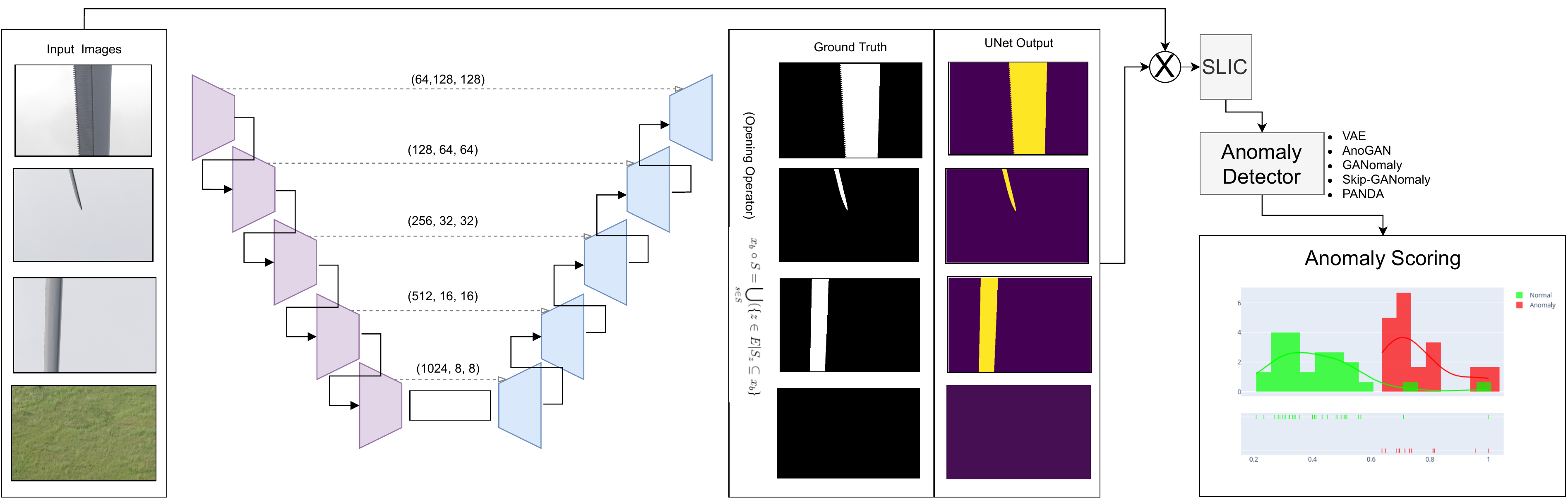}
\caption{Outline of BladeNet Architecture. \textbf{left:} UNet segmentation module which returns the instance segmentation mask of blades in the input images. \textbf{right:} Super pixel and anomaly detection pipeline.}
\label{fig:architecture}
\end{figure*}

\subsection{Semi-supervised Anomaly Detection}\label{subsec:Semi-supervised_AD}
Anomaly detection is the task of recognising artifacts in given data which deviate significantly from normality. Due to the open-bound distribution of anomalous data, it is impossible to account for all forms in which an anomaly may present. Semi-supervised anomaly detection methods \cite{Schlegl2019,Baur2018,Vu2019,Akcay2018,Akcay2019,Barker2021} overcome this by training solely across the benign/non-anomalous data. This allows the models to learn bespoke representations that maps well to benign data, but causes large residual values for anomalous regions.

AnoGAN \cite{Schlegl2019} is the first generative semi-supervised method of anomaly detection. This method utilises a Generative Adversarial Network (GAN) \cite{Goodfellow2014} based architecture which closely approximates the true distribution of the normal data however, it experiences slow inference time due to the computational complexity of remapping to the latent vector space. EGBAD \cite{zenati2018} addresses this inefficiency by simultaneously mapping from image space to latent space using BiGAN \cite{Donahue2019} which results in faster inference times. GANomaly \cite{Akcay2018} better approximates the true distribution by jointly training a generator module together with a secondary encoder in order to re-map the generated samples into a second latent space which is then used to better learn the original latent priors. Generative methods have been greatly improved by implementing residual skip-connections \cite{Akcay2019}. PANDA \cite{Barker2021} utilises a dual-feature extraction method and feature merging together with a bespoke fine-grained classifier to better account for subtle differences between normal and anomalous data.

\subsection{Wind Turbine Blade Surface Defect Detection} \label{subsec:fault_detection_WTB}

Several methods of visual surface fault detection on wind turbine blades using machine-learning based methods have been proposed \cite{Wang2017,Denhof2019,Reddy2019,Shihavuddin2019}. \cite{Wang2017} detects surface cracks of wind turbine blades using data obtained from an aerial drone. Performance is poor however, due to the use of Haar features which are static, manually determined kernels which exhibit poor rotational invariance.

Recent works \cite{Denhof2019,Reddy2019} utilise CNN-based classifiers which greatly improve the classification capability. \cite{Shihavuddin2019} also present their work on deep learning methods applied to drone inspection footage of wind turbine blades. In this work, they utilise object detection using the Faster R-CNN architecture \cite{FasterRCNNRen2015} to detect defined anomalous regions within images. Faster R-CNN however, relies heavily on manual annotation of objects, in this case anomalous parts and has a set number of discrete classes. Four classes are included in the study by Shihavuddin \textit{et al}: leading edge erosion, vortex generator panel (VG), VG with missing teeth, and lightning receptor. 

Methods by \cite{Wang2017,Denhof2019,Reddy2019,Shihavuddin2019} are all supervised methods. Due to having few, discrete classes for an open-set anomaly detection problem, the method outlined in this prior work cannot generalise to detecting the varying nature of real-world blade damage. In contrast, our BladeNet approach provides unsupervised blade detection as well as semi-supervised anomaly detection which solely requires healthy blade data which would be trivial to obtain from factory-new blades. From this, BladeNet can infer and generalise to detect any future anomalies which may present on any blade surface.
\section{Approach}  \label{sec:proposal}
\begin{figure*}[htb!]
\centering
\includegraphics[width=\linewidth]{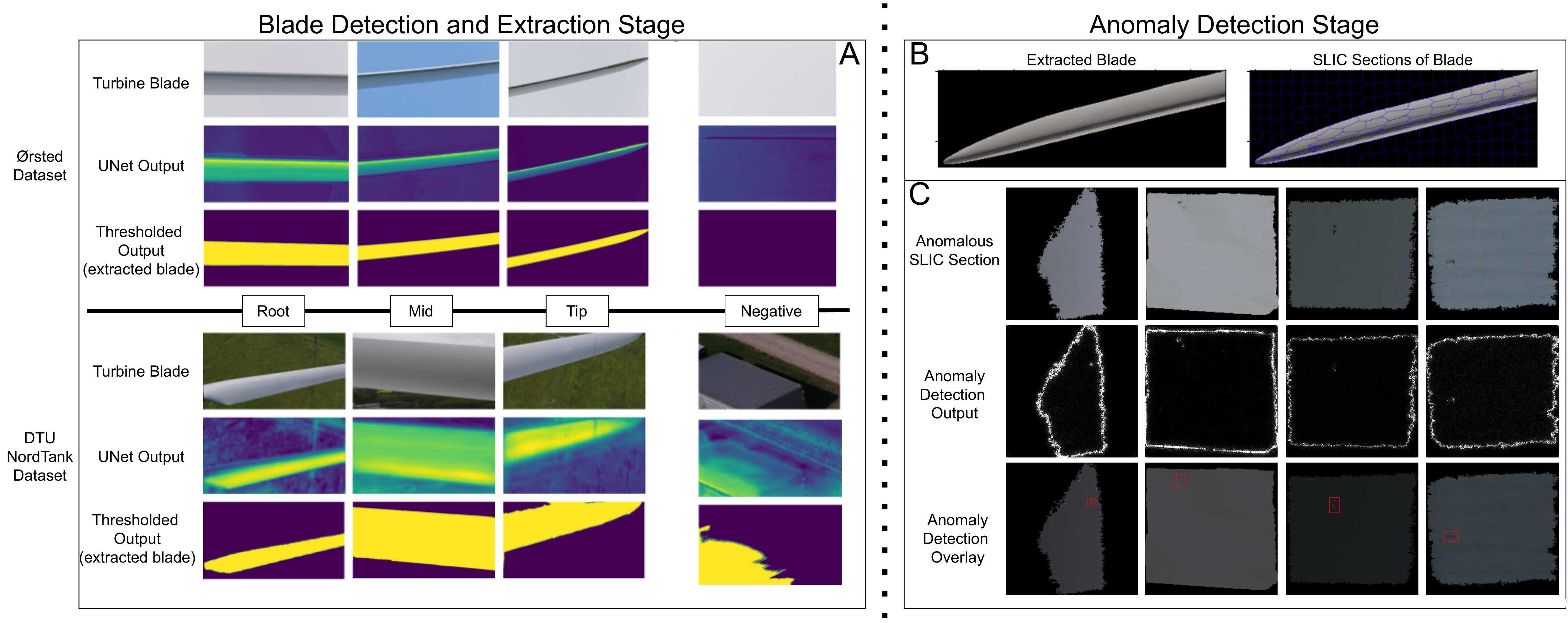}
\caption{ The dual process of detecting surface fault anomalies using BladeNet. \textbf{A) top:} data obtained from Ørsted turbine blade inspection, \textbf{bottom:} DTU NordTank turbine blade inspection data. \textbf{B) left:} Extracted blade using the UNet detector. \textbf{right:} The boundaries of SLIC sections processed over the extracted turbine blade.  \textbf{C)} The anomaly detection of anomalous super-pixel sections using the PANDA \cite{Barker2021} semi-supervised anomaly detection algorithm.}
\label{fig:overview}
\end{figure*}

The BladeNet dual pipeline is outlined in Figure \ref{fig:architecture}, which comprises of operations: blade detection and extraction (Section \ref{subsec:bdae} and Figure \ref{fig:overview}:A) to extract the foreground turbine blade from the background. Extracted blades are then subsequently processed with Simple Linear Iterative Clustering (SLIC) \cite{SLICAchanta2012} (Section \ref{slic} and Figure \ref{fig:overview}:B) to generate super-pixel clusters which are used to train a semi-supervised anomaly detection approach (Section \ref{subsec:ad} and Figure \ref{fig:overview}:C).

\subsection{Unsupervised Blade Detection and Extraction} \label{subsec:bdae}
BladeNet requires accurate blade extraction due to the semi-supervised manner in which the anomaly detection is conducted (Section \ref{subsec:ad}). If background is introduced, or parts of a blade are missing from the non-anomalous training data, the semi-supervised anomaly detection methods \cite{Schlegl2019,Akcay2018,Akcay2019,Barker2021} will not learn adequate, clean representations of non-anomalous blade parts.

\begin{figure*}[h!]
\centering
\includegraphics[width=\linewidth]{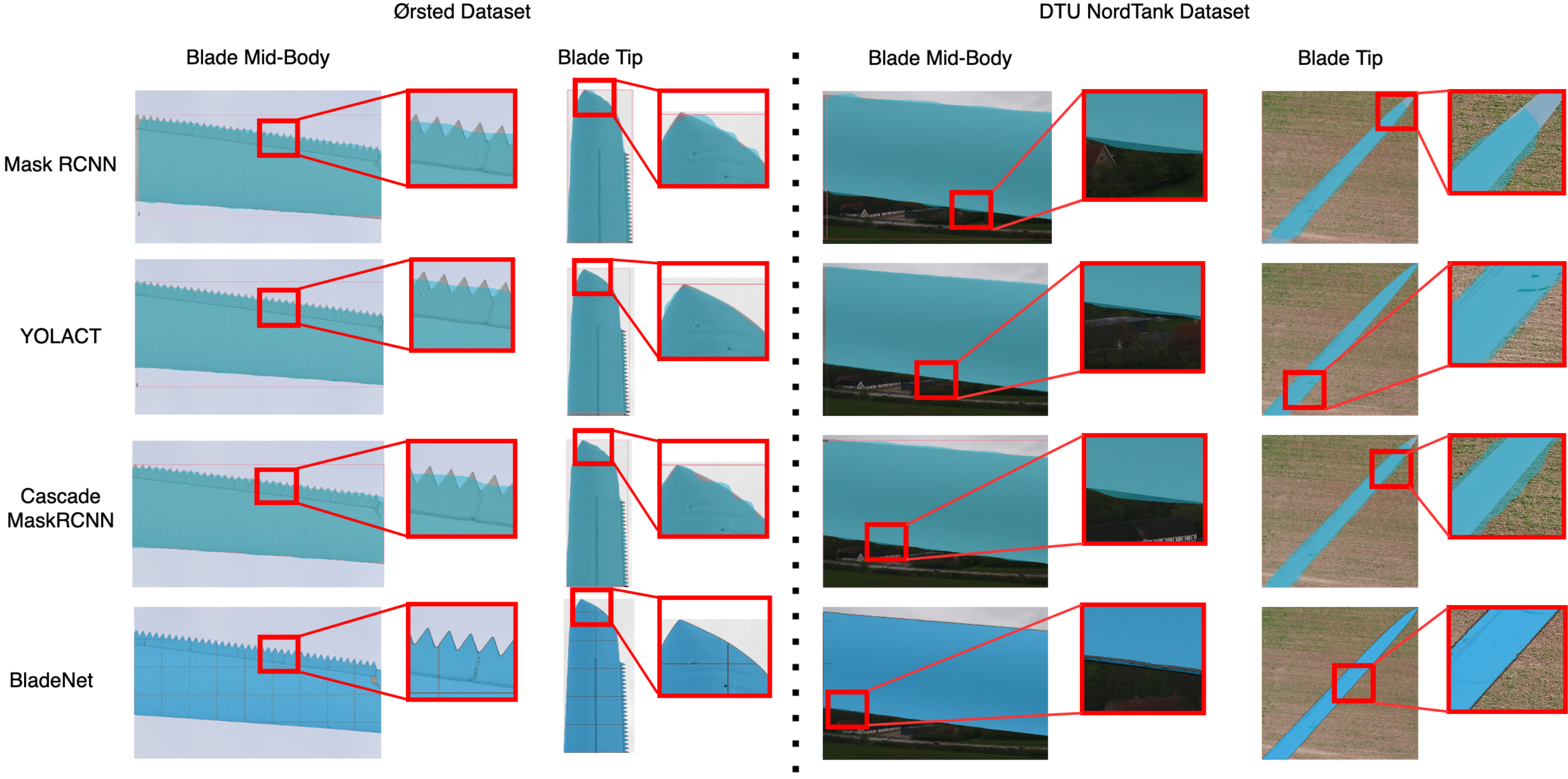}
\caption{Instance segmentation mask quality comparison between Mask R-CNN \cite{MaskRCNNHe2017}, YOLACT \cite{YOLACTBolya2019}, Cascade Mask R-CNN \cite{CascadeMaskRCNNCai2018} and BladeNet.}
\label{fig:maskrcnn_bladenet}
\end{figure*}

When detecting large objects such as turbine blades in high-resolution (6720 $\times$ 4480) drone imagery, conventional instance segmentation models \cite{MaskRCNNHe2017,YOLACTBolya2019,CascadeMaskRCNNCai2018} output masks which appear wavy when placed over the object in the original image. This is due to resizing of the predicted mask from a small resolution up to the full image resolution which exacerbates the loose fit of the mask boundary due to the exaggeration of edges in the small mask. Detection methods also use discrete polygon annotations for objects which under-sample and can fail to capture true curves with enough precision. Our experiments show qualitatively (Figure \ref{fig:maskrcnn_bladenet}) that the masks of Mask R-CNN, YOLACT and Cascade Mask R-CNN all exhibit oscillating detection boundaries around the straight edges of the blades as well as failing to capture important sections of the blade such as the tip and triangular edges of the blades which have the potential to feature anomalies. 

Our approach extracts turbine blade parts from a given image and discards background and unwanted artifacts by utilising a Fully Convolutional (FCN) U-Net \cite{UNetRonneberger2015} architecture for one-class instance segmentation. This architecture is outlined in Figure \ref{fig:architecture}. Five convolutional encoders are used to encode images to a latent representation of shape $1024 \times 8 \times 8$. Five convolutional transpose layers connected in series as well as with residual connections to their encoder counterparts are then used to decode to a 1-channel mask outlining where a blade is present in a given image. This process is illustrated in Figure \ref{fig:blade_detections} in which fixed image patches are taken from the original image (Figure \ref{fig:blade_detections}: 1 and 2) and then inputted into the U-Net module to produce an attention mask (Figure \ref{fig:blade_detections}: 3). A threshold is then applied to this output, producing a clean segmentation mask (Figure \ref{fig:blade_detections}: 4) of turbine blade parts in the original patch.

To create `pseudo ground truth' for our model, we utilise morphology operators and negative example sampling. Using our Ørsted turbine blade inspection dataset $X_{b}$; for each $x_{b} \in X_{b}$ where $x_{b} \in \mathbb{R} ^{B\times 3 \times H \times W}$, the Opening Morphology Operator $x_{b} \circ S = \bigcup_{s \in S} (\{z \in E | S_{z} \subseteq x_{b}\})$ as a combination of erosion $x_{b} \circleddash S$ followed by dilation $x_{b} \oplus  S$ provides pseudo ground truth for $\forall{X_{b}}$ which closely approximates the true edges of the wind turbine blades in $X_{b}$. Negative class examples $x_{n} \notin X_{b}$ consisting of images of sky and ground are introduced during training with a ground truth tensor of zeros of shape $\mathbb{R} ^{B\times 3 \times H \times W}$, indicative of no blade presence in the image. An example of the negative sampling is given in Figure \ref{fig:overview}:A, showing only sky. In this way, BladeNet learns what it must pay attention to, and ignore in a given scene.

\subsection{Superpixel Extraction} \label{slic}
In this work, we implement Simple Linear Iterative Clustering (SLIC) \cite{SLICAchanta2012} for generating sub-region patches of the full blade rather than using conventional sliding window patches. 

Approximately $n$ clusters of neighbouring pixels are generated by stepping over an image of resolution $N = X \times Y$ with an interval $I = |\sqrt{\frac{N}{|n|}}|$ and taking a set of $|n|$ centre points $C = \forall n \in I, \left \{ x_{n},y_{n}\right \}$. Each centre $c_{n} \in C$ is refined by taking the best matching pixels from the neighbourhood of $2S^{2} < X \times Y | S \in \mathbb{N}$ surrounding pixels utilising euclidean distance upon both the pixel colour vector ($lab$) and the pixel coordinates as: $D_s = \sqrt{(l_{n} - l_{i})^2+(a_{n}-a_{i})^2 + (b_{n} - b_{i})^2} + \frac{m}{S}\sqrt{(x_{n} - x_{i})^2+(y_{n} + y_{i})^2}$ where $m$ is the spatial proximity factor of the method.

SLIC patches contain pixels which share visual characteristics to other pixels belonging to the same super-pixel. Super-pixels increase the likelihood that an anomalous region in the image, or key region of interest for a given blade will not be situated across the edge of two neighbouring patches. If an anomalous region is split across two patches, then it not only decreases the size of region by the size of the overlap, but the edge of the patch restricts the features of the area surrounding the anomalous region to only the edge of the image hence the model will not be fully utilising the spatial information of the anomalous region.

\subsection{Anomaly Detection} \label{subsec:ad}

Semi-supervised anomaly detection is performed by using those super-pixels which have no visible defects featured to train a generative model to map to a representation manifold such that when a visual defect presents itself, the representation will differ from normality and as such, the presented example will be flagged as anomalous by the model.

In this work, we utilise a number of self-supervised anomaly detection algorithms \cite{Schlegl2017,Akcay2018,Akcay2019,Barker2021} to evaluate which one is best suited to this task of detecting surface faults in composite blade materials.
\section{Experimental Setup} \label{sec:eva}
We evaluate the performance of the BladeNet architecture by individually comparing each component. We start with the blade detection and extraction (Section: \ref{bde}) and then the anomaly detection of anomalous regions on the blade surfaces (Section: \ref{adsd}).

The two datasets used in this paper are Ørsted turbine blade inspection dataset and the DTU NordTank blade inspection dataset. The Ørsted turbine blade inspection dataset consists of drone inspection imagery of offshore wind turbine blades from the Hornsea 1 wind farm. It contains 2637 images of resolution 6720 $\times$ 4480 of offshore turbine blades from varying perspectives in differing weather and backdrop. The DTU NordTank dataset is supplied by \cite{Shihavuddin2019} and contains drone imagery from 1170 onshore wind turbines. In both datasets we use a 20:80 split for testing and training data respectively.

We evaluate BladeNet against established benchmark methods. We train our detection method solely across the Ørsted turbine blade inspection dataset together with negative image samples. After training, we infer across the the DTU NordTank dataset using the same learned model parameters to demonstrate the robustness of our approach.

All training was performed on a Titan X GPU. `Binary Cross Entropy (BCE) with logits' loss with a learning rate of 0.001 was utilised for the U-Net blade detector along with RMS Prop optimiser with weight decay of $1e^{-8}$ and momentum of 0.9. Image scaling by 0.2 was also performed to preserve memory usage with a batch size of 10. Augmentation of rotation (degrees 90, 180, 270), flipping with probability 0.5, and random crop were used during training.

\section{Evaluation} \label{sec:eva}

%%%%%%%%%%%%%%%%%%%%%%%%%%%%%%%%%%%%%%%%%%%%%%%%%%%%%%%%%%%

% \begin{table*}[htb!]

% \centering
% \caption{Average precision (AP) at IoU = 0.5, number of parameters in Millions.}
% \label{tab:object_detection}
% \addtolength{\tabcolsep}{-5.0pt} 

% \resizebox{450pt}{!}{
% \begin{tabular}{lc|c|c|c|c|c|c|}
% \cline{3-8}
%  & \multicolumn{1}{l|}{} & \multicolumn{2}{l|}{\textbf{Ørsted Dataset}} & \multicolumn{1}{l|}{\textbf{ DTU NordTank Dataset}} & \multicolumn{2}{l|}{\textbf{Ørsted Model $\rightarrow$  DTU NordTank Dataset}} & \multicolumn{1}{l|}{\textbf{DTU NordTank Model $\rightarrow$ Ørsted Dataset}} \\ \cline{2-8} 
% \multicolumn{1}{l|}{} & \textbf{Params} & \textbf{AP} & \textbf{Time (ms)} & \textbf{AP} & \textbf{AP} & {\textbf{Time(ms)}} & \textbf{AP} \\ \hline
% \multicolumn{1}{|l|}{Mask R-CNN} & 43.9 & 0.983 & 590.36 & \textbf{0.958} & 0.005 & 537.31 & 0.235 \\ \hline
% \multicolumn{1}{|l|}{YOLACT}  & 34.7 & 0.983 & 549.06 & 0.933 & 0.023 & 478.04 & 0.175 \\ \hline
% \multicolumn{1}{|l|}{Cascade Mask R-CNN}  & 77 & 0.985 & \textbf{520.12} & 0.948 & 0.002 & \textbf{314.61} & \textbf{0.246} \\ \hline
% \multicolumn{1}{|l|}{\textbf{BladeNet}} & \textbf{17.3} & \textbf{0.995} & 3439.21 & - & \textbf{0.223} & 1791.43 & - \\ \hline
% \end{tabular}
% }
% \addtolength{\tabcolsep}{1pt} 

% \label{tab:map}
% \end{table*}
%%%%%%%%%%%%%%%%%%%%%%%%%%%%%%%%%%%%%%%%%%%%%%%%%%%%%%%%%%%

\begin{table*}[htb!]

\centering
\caption{Average precision (AP) at IoU = 0.5, number of parameters in Millions.}
\label{tab:object_detection}
\addtolength{\tabcolsep}{-5.0pt} 

\begin{tabular}{lc|c|c|c|c|}
\cline{3-6}
 & \multicolumn{1}{l|}{} 
 & \multicolumn{2}{l|}{\textbf{Ørsted Dataset}} 
 & \multicolumn{2}{l|}{\textbf{Ørsted Model $\rightarrow$  DTU NordTank Dataset}} 
\\ \cline{2-6} 
\multicolumn{1}{l|}{} & \textbf{Params} & \textbf{AP} & \textbf{Time (ms)} & \textbf{AP} & {\textbf{Time(ms)}}\\ \hline
\multicolumn{1}{|l|}{Mask R-CNN} & 43.9 & 0.983 & 590.36 & 0.005 & 537.31 \\ \hline
\multicolumn{1}{|l|}{YOLACT}  & 34.7 & 0.983 & 549.06 & 0.023 & 478.04 \\ \hline
\multicolumn{1}{|l|}{Cascade Mask R-CNN}  & 77 & 0.985 & \textbf{520.12} & 0.002 & \textbf{314.61}  \\ \hline
\multicolumn{1}{|l|}{\textbf{BladeNet}} & \textbf{17.3} & \textbf{0.995} & 3439.21 & \textbf{0.223} & 1791.43 \\ \hline
\end{tabular}

\addtolength{\tabcolsep}{1pt} 

\label{tab:map}
\end{table*}

\subsection{Blade Detection and Extraction} \label{bde}

The quantitative performance outlined in Table I shows that Mask R-CNN performed equally in Average Precision (AP) with YOLACT at 0.983 across the Ørsted dataset however, YOLACT obtained a greater AP value of 0.023 on the transfer to the DTU NordTank dataset. Cascade Mask R-CNN surpassed the performance of YOLACT across the Ørsted dataset and achieved the best time efficiency of 520.12 ms of all models in the study, but performs worse than Mask R-CNN across the DTU NordTank dataset with AP of 0.002. Our method, BladeNet performs the best quantitatively, obtaining an AP of 0.995, 0.1 higher than the next best performing (Cascade Mask R-CNN) and an AP of 0.223 on the transfer DTU NordTank dataset, far out-performing all prior methods bespoke to the task of object detection.

\begin{figure}[t!]
\centering
\includegraphics[width=\linewidth]{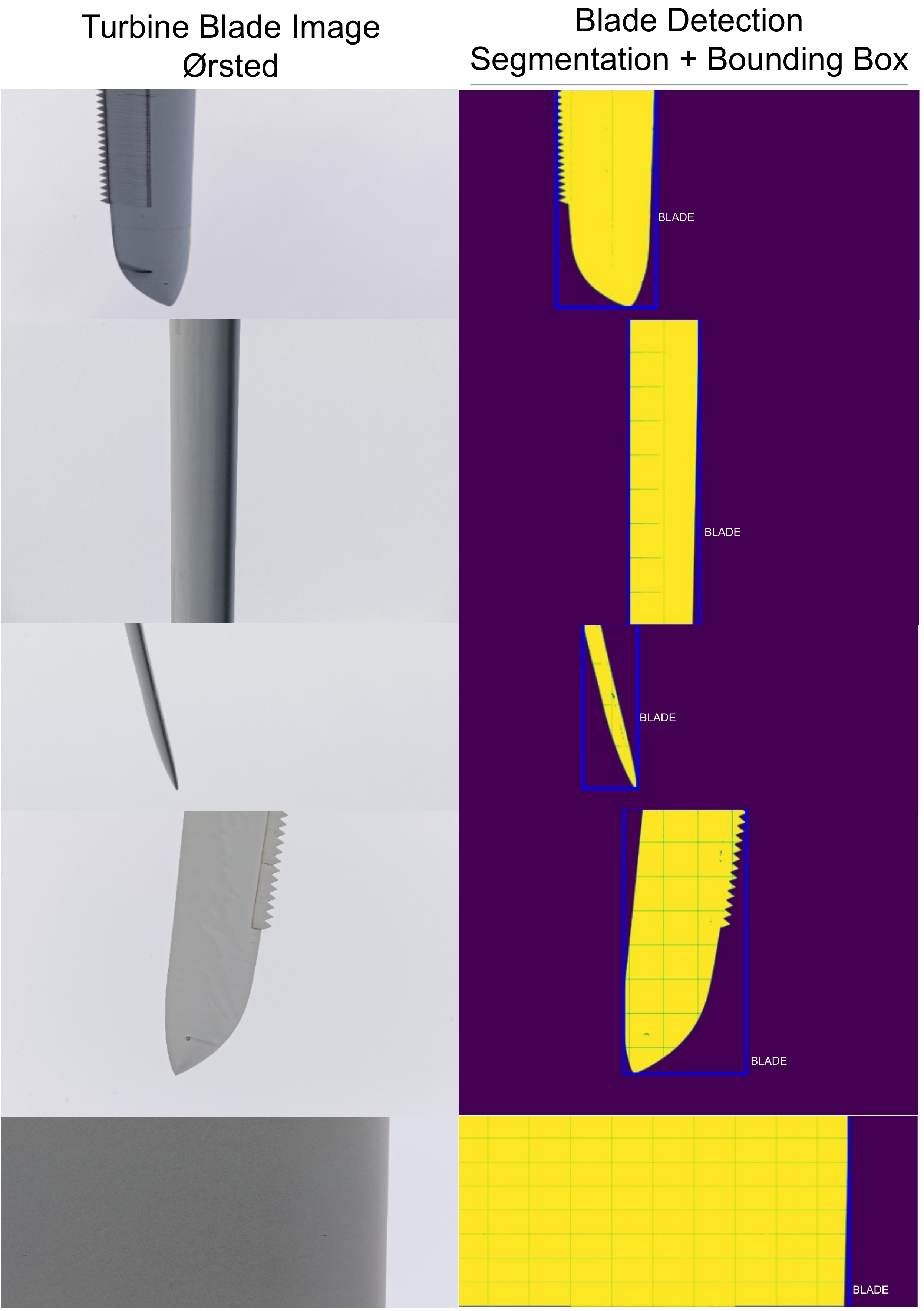}
\caption{Examples of high accuracy instance segmentation and bounding box prediction of Ørsted turbine blades using BladeNet.}
\label{fig:bounding_box}
\end{figure}

BladeNet produces clean and sharp masks which fit the blades closely and manage to detect the sharp triangular parts of the mid-body blade and the blade tip with high precision. These masks can be seen in Figure \ref{fig:maskrcnn_bladenet} when zooming in on the edge of the mask predictions, BladeNet remains tight with the true edge of the blade. Figure \ref{fig:bounding_box} further shows this capability of BladeNet at detecting numerous Ørsted turbine blade parts from different poses and angles with high accuracy. The other such methods such as Mask R-CNN and Cascade Mask R-CNN outlined in Figure \ref{fig:maskrcnn_bladenet}, fit the turbine blades poorly; missing out important sections of the blade edge which are prone to anomalies (edge erosion) in their mask predictions. Using these methods would enable null-categorisation of such parts of the blade and hence impose false-negative error due to anomalous regions going undetected.

In Figure \ref{fig:maskrcnn_bladenet}:A, detection across both the Ørsted and DTU NordTank dataset can be seen together with the respective attention mask for the blades. It is interesting that for the negative sample on the DTU NordTank dataset, BladeNet mistakenly predicts that the metal corrugated roof of the building is a turbine blade due to the colour and straight edges of the roof, resembling that of a turbine blade.

\subsection{Anomaly Detection of Surface Defects} \label{adsd}
We include a quantitative study of Semi-Supervised anomaly detection approaches over the extracted SLIC super-pixel data of turbine blades. It can be seen in Table II that PANDA gains the highest Area Under Curve (AUC) value at 0.639 and obtains a tight 95\% Confidence Interval (CI) between 0.631 and 0.648. This is comparatively close to the performance of Skip-GANomaly which obtains 0.631 however these models suffer from slower relative inference time compared to that of the Variational Autoencoder which obtained 0.625 (0.14 lower than PANDA), but only took 8.61 milliseconds compared with PANDA at 50.3. AnoGAN exhibits sluggish inference speed of over 300ms for prediction and obtains the lowest AUC value of 0.611 however, the 95\% CI is similar to that of the VAE architecture.

The qualitative results of PANDA across the SLIC super pixels of the blade data can be seen in Figure \ref{fig:overview}:C. Note that the edge of the blade is considered anomalous due to the random shape that SLIC superpixels pose however, once thresholded, the anomalous regions can be seen clearly when overlayed on top of the original blade super-pixel. The localisation manages to locate the anomalous regions within the super-pixel.

\begin{table}[htb!]
\caption{Area Under Curve (AUC) of ROC curve, inference time per image in Milliseconds (I/t(ms)) across semi-supervised anomaly detection methods.}
\addtolength{\tabcolsep}{-3.0pt} 

\begin{tabular}{|l|l|l|l|}
\hline
\textbf{Model} & \multicolumn{1}{c|}{\textbf{AUC}} & \multicolumn{1}{c|}{\textbf{\begin{tabular}[c]{@{}c@{}}95\% CI \\ (AUC)\end{tabular}}} & \multicolumn{1}{c|}{\textbf{I/t/(ms)}} \\ \hline
VAE           & 0.625                            & 0.609\textless{}x\textless{}0.626                                                     & \textbf{8.61}                         \\ \hline
AnoGAN        & 0.611                            & 0.608\textless{}x\textless{}0.625                                                     & 302                                   \\ \hline
GANomaly       & 0.628                            & 0.61\textless{}x\textless{}0.634                                                      & 48.36                                 \\ \hline
Skip-GANomaly  & 0.631                            & 0.621\textless{}x\textless{}0.636                                                     & 97.21                                 \\ \hline
\textbf{PANDA} & \textbf{0.639}                   & \textbf{0.631\textless{}x\textless{}0.648}                                            & 50.3                                  \\ \hline
\end{tabular}
\label{tab:anomaly}
\addtolength{\tabcolsep}{1.0pt} 

\end{table}

\clearpage

\section{Conclusion} \label{sec:conclusion}
In this work we propose BladeNet, an application-based approach for detecting surface-fault anomalies on composite material-constructed wind turbine blades using drone imagery. BladeNet utilises an instance-segmentation method of blade extraction which is far more precise at fitting the blade edges than conventional object detection models both qualititively and quantitatively obtaining a Average Precision (AP) of 0.995 together with a suite of semi-supervised generative anomaly detection methods across extracted SLIC super-pixel blade parts to detect anomalies with an AUC of 0.639. We hope that this work can aid engineers and wind farm inspectors to detect surface faults of composite wind turbines.

%%%%%%%%%%%%%%%%%%%%%%%%%%%%%%%%%%%%%%%%%%%%%%%%%%%%%%%%%%%%%%%%

%\vfill
\subsection{\uppercase{Acknowledgements}}

Thank you to EPSRC and Ørsted for funding support towards this work.

% References
\bibliographystyle{ref/apalike}
{\small
\bibliography{ref/example,ref/example2}}

\end{document}